\RequirePackage{fix-cm}
\documentclass[twocolumn]{svjour3}          % twocolumn
\smartqed  % flush right qed marks, e.g. at end of proof
\usepackage{graphicx,amsmath,amssymb,amsfonts, cite}
\usepackage[round]{natbib}
\usepackage{hyperref}
\begin{document}

\title{Variational Inference to Measure Model Uncertainty in Deep Neural Networks \thanks{The results presented in this article have been obtained as part of the master thesis of Konstantin Posch in 2017. The master thesis has been peer reviewed and accepted by the thesis committee according to common University guidelines and standards on October 29th, 2017. It has been made available through the library of the University of Klagenfurt since then.  The results of the Master thesis have been presented at the 23rd Int. Meeting of Young Statisticians in Balatonfüred (Hungary) on October 14, 2018. A version of this article has been submitted to Springer International Journal of Computer Vision on May 7th, 2018 and is currently still under review. We have since been notified about another article published on Arxiv on November 14th, 2018 (\href{https://arxiv.org/pdf/1806.05978.pdf}{https://arxiv.org/pdf/1806.05978.pdf}) using essentially the same approach as we did, i.e. with variances of the variational distribution specified as multiples of the corresponding squared expectation values. However in their approach, different multiplication factors for each network weight are assumed whereas we only use one factor for all weights of one particular layer.}}



\author{Konstantin Posch*  \thanks{*both authors contributed equally to this work}       \and
        Jan Steinbrener* \and
	J\"urgen Pilz %etc.
}


\institute{K. Posch \at
              Alpen-Adria Universit\"at, Klagenfurt am W\"orthersee, Austria \\
              \email{koposch@edu.aau.at}           %  \\
           \and
          J. Steinbrener \at 
	CTR Carinthian Tech Research AG, Villach, Austria \\
	\email{jan.steinbrener@ctr.at}
	\and
	 J. Pilz \at 
	Alpen-Adria Universit\"at, Klagenfurt am W\"orthersee, Austria \\
	\email{juergen.pilz@aau.at}
}

\date{Received: date / Accepted: date}

\maketitle

\begin{abstract}
We present a novel approach for training deep neural networks in a Bayesian way. Classical, i.e. non-Bayesian, deep learning has two major drawbacks both originating from the fact that network parameters are considered to be deterministic. First, model uncertainty cannot be measured thus limiting the use of deep learning in many fields of application and second, training of deep neural networks is often hampered by overfitting. The proposed approach uses variational inference to approximate the intractable a posteriori distribution on basis of a normal prior. The variational density is designed in such a way that the a posteriori uncertainty of the network parameters is represented per network layer and depending on the estimated parameter expectation values. This way, only a few additional parameters need to be optimized compared to a non-Bayesian network. We apply this Bayesian approach to train and test the LeNet architecture on the MNIST dataset. Compared to classical deep learning, the test error is reduced by 15\%. In addition, the trained model contains information about the parameter uncertainty in each layer. We show that this information can be used to calculate credible intervals for the prediction and to optimize the network architecture for a given training data set.
\end{abstract}

\begin{acknowledgements}
This work was performed within the Competence Centre 'ASSIC Austrian Smart Systems Integration  Research Center', co-funded by the Federal Ministries of Transport, Innovation and Technology (bmvit) and Science, Research and Economy (bmwfw) and the Federal Provinces of Carinthia and Styria within the COMET - Competence Centers for Excellent Technologies Programme. This work was also supported by Philips Austria GmbH as part of the cooperation between Philips Austria and CTR within the COMET programme. 
\end{acknowledgements}

\section{Introduction}
\label{intro}
Deep learning has led to series of breakthroughs in many fields of applied machine learning, especially in image classification \citep{Krizhevsky:ImageNet} or natural language processing \citep{Bengio}. In 1989, the universal approximation theorem was proven, which can be summarized that a feed-forward network with one hidden layer can approximate a broad class of functions abitrarily well \citep{hornik}. Currently, it has been shown that for a given bound of the approximation error, deep networks require exponentially less data than shallow ones \citep{Shiyu}. The possible applications of deep neural networks for classification and detection cover a wide range including medical imaging, psychology, automotive, industry, finance and life sciences \citep{gulshanJama2016, GreenspanIEEE2016, LiReliabilityEngin2018, BanerjeeIEEEIVS2017, JozwikFrontiersPsych2017, HeatonASMB2017}.  

Despite its potential and superior accuracy for classification tasks compared to other techniques, dissemination of deep learning into real world applications and services has been limited by a lack of information about model uncertainty. This particularly affects those applications, where wrong decisions based on false classification results could have significant negative or even catastrophic impact such as in self-driving cars, finance or medical applications \citep{Andrew}. In addition, most network architectures today are designed based on trial and error or based on abstract, high level considerations \citep{GoodfellowMITPress2016, AroraProcJMLR2014}. Thus, the process of finding an optimal network architecture for the classification task at hand and given the training data can be cumbersome. Standard deep networks for classification and regression do not represent model uncertainty since network parameters are considered to be deterministic values. Sometimes in classification the probabilities obtained when running the model are falsely interpreted as the confidence of the model, see \citep{Gal:baysianApprox}. Indeed a network can guess randomly while returning a high class probability. Often it is essential to know how sure a network is about a special prediction and not only that it predicts on average quite well. 

Besides the inability of classical deep nets to represent model uncertainty they are prone to overfitting. Modern deep models cover a huge amount of parameters and therefore require a huge amount of labeled training data as well. In many applications, such an amount cannot be provided because of financial or time constraints. To overcome this problem, the deep learning community introduced several probabilistic regularization techniques, such as dropout and dropconnect \citep{Nitish:Dropout, dropconnect}. \cite{Gal:conv} could show that an appropriate application of dropout can be interpreted as training networks in a Bayesian way, by approximating the a posteriori distribution via variational inference.

Both major drawbacks of deep learning, the absence of model uncertainty evaluations and the need of a large amount of training data, are well addressed by using Bayesian statistics. On the one hand Bayesian models are robust to overfitting since 
parameters are not forced to be fixed and on the other hand the uncertainty in the network parameters can directly be translated in uncertainty information for network predictions. Further, Bayesian deep models can help to finally understand why deep learning works. Combining the profound theoretical literature about Bayesian statistics and deep learning will lead to a better understanding and broader acceptance of the technique.

In this study, a new approach for Bayesian deep neural networks based on variational inference is proposed. It is the first approach which treats network layers as units in order to express model uncertainty. Therefore only two uncertainty parameters are introduced per layer which implies that the variational distribution requires only few additional parameters that need to be optimized compared to a non-Bayesian net. Besides good convergence properties, this way of proceeding also provides consolidated information about network uncertainty that can be readily used to optimize network architecture. Introducing too many uncertainty parameters, i.e. one for each network parameter, worsens the convergence properties and results in an excess of information about network uncertainty which is difficult to interpret. 
\section{Theoretical Background and Related Work}
\label{sec:1}
In this section, a short introduction to Bayesian statistics and variational inference is presented. More details can be found in \citep{hintonColt1993, jordanML1999, Blei:VariationalS, Bishop, Gal:baysianApprox, Gal:appendix}. Further, it is summarized how variational inference was applied in the past to train deep neural networks in a Bayesian way, which also sheds some light on the limitations of each approach.

\subsection{Bayesian and Variational Inference}
\label{sec:2}
The theoretical considerations are based on classification tasks in this study. For regression, the theory is quite the same and can be found in the literature recommended above. In Bayesian statistics, network parameters are considered as one large random vector $\underline{W}$. A priori knowledge regarding $\underline{W}$ is expressed in terms of the a priori distribution $p(\underline{w})$. One is interested in updating the knowledge about $\underline{W}$ after observing data $D=\{\underline{y},X\}$, where $X=\{\underline{x}_1,...,\underline{x}_{\beta}\}$ denotes a set of training examples and $\underline{y}=(y_1,...,y_{\beta})^T$ holds the corresponding class labels. Therefore, the a posteriori distribution $p(\underline{w}|\underline{y},X)$ has to be calculated. According to the Bayes' theorem, the corresponding density is
$$p(\underline{w}|\underline{y},X)=\frac{p(\underline{y}|\underline{w},X)p(\underline{w})}{\int p(\underline{y}|\underline{w},X)p(\underline{w})~d\underline{w}}$$
The joint density $p(\underline{y}|\underline{w},X)$ is given by the product of the neural network outputs for all the training examples, following the classical assumptions on stochastic independence and modelling in deep learning. Thus, the only problem in computing $p(\underline{w}|\underline{y},X)$ is the generally intractable integral in the denominator. Variational inference aims at approximating the posterior $p(\underline{w}|\underline{y},X)$ by optimizing a parametric distribution $q_{\underline{\phi}}(\underline{w})$, such that it is most similar to $p(\underline{w}|\underline{y},X)$.

Once the variational distribution is optimized, it can be used for predicting new data and further quantifying uncertainty in predictions. The posterior predictive distribution $p(y^*|\underline{x}^*,\underline{y},X)$ reflects the belief in a class label $y^*$ for a given example $\underline{x}^*$ after observing data $\underline{y},X$:
\begin{align*}
p(y^*|\underline{x}^*,\underline{y},X)&=\int p(y^*,\underline{w}|\underline{x}^*,\underline{y},X)~d\underline{w}\\
&=\int p(y^*|\underline{w},\underline{x}^*,\underline{y},X)p(\underline{w}|\underline{y},X)~d\underline{w}\\
&=\int p(y^*|\underline{w},\underline{x}^*)p(\underline{w}|\underline{y},X)~d\underline{w}
\end{align*}
Replacing the posterior with the variational distribution and further approximating the intractable integral via Monte Carlo integration results in
$$p(y^*|\underline{x}^*,\underline{y},X) \approx \frac{1}{N}\sum_{i=1}^N \underline{f}(\underline{x}^*;\widehat{\underline{w}}_i)_{y^*} ~~~\text{with } \widehat{\underline{w}}_i\underset{i.i.d.}{\sim}q_{\underline{\phi}}(\underline{w}) $$
where $\underline{f}$ denotes the neural network used. Thus, predictions $\widehat{y^*}$ are made  by propagating the object of interest $\underline{x}^*$ multiple times through the network, averaging the resulting probability vectors and choosing the index of the largest element in the resulting mean:
$$\widehat{y^*}=\underset{j}{\operatorname{arg~max}}\frac{1}{N}\sum_{i=1}^N \underline{f}(\underline{x}^*;\widehat{\underline{w}}_i)_{j}$$
To measure model uncertainty, credible intervals can be estimated. The output of a neural net $\underline{f}(\underline{x}^*,\underline{w})_{y^*}$ with fixed parameters equals the probability $p(y^*|\underline{w},\underline{x}^*)$. Further, the variational distribution $q_{\underline{\phi}}(\underline{w})$ reflects the knowledge about the random network parameters $\underline{W}$. The uncertainty in the parameters implies an uncertainty in the neural network output, i.e. the probability of class $y^*$. By sampling from the variational distribution and subsequently calculating the empirical  $\frac{\alpha}{2}$ and $1-\frac{\alpha}{2}$ quantiles of the corresponding network outputs $\underline{f}(\underline{x}^*;\widehat{\underline{w}}_1)_{y^*},...,\underline{f}(\underline{x}^*;\widehat{\underline{w}}_N)_{y^*}$ an estimate of the $1-\alpha$ credible interval for the probability of $y^*$ is found. Note that neural network outputs are uniquely determined for fixed $\underline{w}$ and therefore sampling from the output distribution is equivalent to sampling from the parameters.

So far it has not been mentioned how the variational distribution is optimized in order to approximate the posterior $p(\underline{w}|\underline{y},X)$. This can be accomplished by minimizeing the Kullback-Leibler divergence \textit{(KL-divergence)} $D_{KL}(q_{\underline{\phi}}(\underline{w})||p(\underline{w}|\underline{y},X))$ between the variational density and the posterior. It is defined as:
\begin{align*}
D_{KL}(q_{\underline{\phi}}(\underline{w})||p(\underline{w}|\underline{y},X))&:=\mathbb{E}_{q_{\underline{\phi}}(\underline{w})}\left(\operatorname{ln}\frac{q_{\underline{\phi}}(\underline{w})}{p(\underline{w}|\underline{y},X)}\right)\\
&=\int\operatorname{ln}\frac{q_{\underline{\phi}}(\underline{w})}{p(\underline{w}|\underline{y},X)}q_{\underline{\phi}}(\underline{w})~d\underline{w}
\end{align*}
The KL-divergence is not really a distance measure since it is asymmetric and the triangle inequality does not hold. Nevertheless, it is often used to measure the distance between two probability distributions, and as long as only two distributions are of interest it does not matter that the triangle inequality is violated. Obviously, $D_{KL}(q_{\underline{\phi}}(\underline{w})||p(\underline{w}|\underline{y},X))$ cannot be minimized directly since the a posteriori distribution is unknown. However, minimizing  $D_{KL}(q_{\underline{\phi}}(\underline{w})||p(\underline{w}|\underline{y},X))$ is equivalent to minimizing the negative log evidence lower bound $L_{VI}$ \citep{Bishop}, which is given by:
\begin{align*}
L_{VI}&=-\int q_{\underline{\phi}}(\underline{w})\operatorname{ln}p(\underline{y}|\underline{w},X)~d\underline{w}+D_{KL}(q_{\underline{\phi}}(\underline{w})||p(\underline{w}))\\
&=-\mathbb{E}_{q_{\underline{\phi}}(\underline{w})}\left[\operatorname{ln}\prod\limits_{i=1}^{\beta}p(y_i|\underline{w}, \underline{x}_i)\right]+D_{KL}(q_{\underline{\phi}}(\underline{w})||p(\underline{w}))\\
&=-\sum\limits_{i=1}^{\beta}\left[\mathbb{E}_{q_{\underline{\phi}(\underline{w})}}\left(\operatorname{ln} p(y_i|\underline{w}, \underline{x}_i)\right)\right]+D_{KL}(q_{\underline{\phi}}(\underline{w})||p(\underline{w}))
\end{align*}
$L_{VI}$ includes the KL-divergence between the variational distribution and the well known prior. The unknown expectation value can be approximated via Monte Carlo integration. Inspired by stochastic gradient descent the integration takes place with just one sample, but a new sample is drawn in each iteration of the optimization procedure used to minimize $L_{VI}$. The re-sampling guarantees that a sufficient amount of samples is drawn, whilst using merely one sample saves memory. According to these considerations, the objective function in the $k$-th iteration of the optimization is given by:
$$L_{VI}^{k}=-\sum\limits_{i=1}^{\beta}\operatorname{ln}\underline{f}(\underline{x}_i;\widehat{\underline{w}}_k)_{y_i}+D_{KL}(q_{\underline{\phi}}(\underline{w})||p(\underline{w}))$$
If one wants to use mini-batch gradient descent, the KL-divergence has to be rescaled by the factor $\frac{m}{•\beta}$, where $m$ denotes the number of examples one mini-batch holds. This ensures that the divergence does not get too much weight.

Summing up, training neural networks in a Bayesian way via variational inference is quite similar to frequentist training. The $L^{2}$-norm regularization used in classical deep learning is replaced by punishing deviations from the a priori distribution. The same error function as in non-Bayesian deep learning is used with the crucial difference that the network parameters are drawn from the variational distribution during training.

\subsection{Related Work}

There is a large number of possibilities of how to choose the variational distribution. \cite{Gal:conv} have shown how classical Bernoulli dropout can be used to define the approximating function. The network biases are assumed to be deterministic for simplicity, whilst the network weights are defined to be random according to dropout. Indeed, randomly dropping a neuron in layer $i-1$ is equivalent to dropping all weights in layer $i$ which represent connections to this one neuron. In order to calculate the KL-divergence to a standard normal prior, network weights are assumed to follow a mixture of two Gaussians. Note that the KL-divergence between a discrete and a continuous distribution would diverge to infinity. Both Gaussians are defined to have a variance that is negligibly small, such that more or less only two values \textit{(zero and a variational parameter to be optimized)} are taken. Finally, the KL-divergence is given by the $L^2$-norm of the neural network weights. Therefore, neural nets can be learned in a Bayesian way by merely applying dropout after each weight layer except the last one. Experiments have shown that this approach results in a very good accuracy at the MNIST dataset of handwritten digits \citep{Gal:conv}.

\cite{reparam} used a normal distribution with a diagonal covariance matrix as variational distribution. The approach was evaluated with the LeNet architecture \citep{lenet} and the MNIST dataset. While the approach was shown to work in principle, a wider application is hampered by the fact that the number of parameters to be optimized is doubled \textit{(one variance term for each expectation value)} which complicates training and makes it computationally significantly more expensive. \cite{LouizosArxive2016} introduced a variational distribution that in contrast to the distribution of Blundell et al. does not treat each network parameter independently. In particular, they used a probability distribution on random matrices. Thus, they could reduce the number of variance-related parameters, but to a number which nonetheless is significantly higher than in the frequentist approach.
Further approaches with similar restrictions are described in \cite{Kingma} and \cite{Graves}.

It should be mentioned that variational Bayes is just a specific case of local $\alpha$-divergence minimization. The $\alpha$-divergence \citep{AmariSpringer1985} between two densities $p(\underline{w})$ and $q(\underline{w})$ is defined as
$$D_{\alpha}(p(\underline{w})||q(\underline{w}))=\frac{1}{\alpha(1-\alpha)}\left(1-\int p(\underline{w})^{\alpha}q(\underline{w})^{1-\alpha}~d\underline{w}\right)$$
such that $D_{\alpha}(p(\underline{w})||q(\underline{w}))$ converges to the Kullback-Leibler divergence $D_{KL}(q(\underline{w})||p(\underline{w}))$ for $\alpha\rightarrow 0$. \\\cite{LobatoArxiv2015} have shown that the optimal setting for $\alpha$ is task specific and that a nonstandard stetting $\alpha\neq 0$ can produce better prediction results. \cite{LiArxiv2017} continued the work of Hern\'{a}ndez-Lobato et al.. According to them, variational inference can underestimate model uncertainty and $\alpha$-divergences are able to avoid the underestimation. In particular they propose a simple inference technique based on a re-parametrisation of the $\alpha$-divergence objectives and dropout.
However, our work does not focus on finding an optimal choice for $\alpha$. It tries to propose a good and reasonable approximating distribution. The proposed distribution can then be used with any setting of $\alpha$, but this is left for further research. Despite our restriction to $\alpha=0$, our best model on the MNIST dataset (see section 4.2) shows an accuracy which is slightly better than the accuracy of the best model obtained in \cite{LiArxiv2017}.

\section{Our Work}

In this work, we propose a variational distribution with the aim to satisfy the following requirements:
\begin{itemize}
\item uncertainty of predictions can be reliably measured
\item the number of parameters to be optimized does not differ significantly from the non-Bayesian case
\item model uncertainty information is given
\end{itemize}

So far, there is no variational distribution recommended in previous work which satisfies all the requirements. The approach of \cite{Gal:conv} satisfies the second requirement since the number of parameters to be optimized is exactly the same as in frequentist deep learning. In addition, uncertainty of predictions (the first requirement) can be measured, but it is questionable how well their distribution can approximate the complex a posteriori distribution and therefore how reliable the uncertainty information is. The third point is clearly not satisfied. The approach of \cite{reparam} satisfies the first requirement. The second requirement is violated since the amount of parameters is doubled. This also makes it hard to evaluate the model uncertainty (third requirement). Deep architectures typically hold millions of parameters. Computing individual uncertainties for each parameter makes it difficult to extract useful information for the whole network.

In order to satisfy all three requirements, our approach expresses model uncertainty layer-wise with respect to the parameter expectation values. Therefore, only two uncertainty parameters are introduced per layer which leads to good convergence properties. Interpreting network layers as units in terms of uncertainty enables a quick understanding of the model considered. Strong uncertainty in some layers indicate that the architectural designs of them might not be optimal and should be reconsidered. Thus in addition to allow for the uncertainty of the predictions to be measured, our approach also helps for finetuning network architectures. At first glance, the requirement number one -reliable uncertainty information of predictions- might not be satisfied as well by our approach as by \cite{reparam}. For sure, the more complex distribution proposed by them should theoretically lead to a better approximation of the posterior and therefore more reliable uncertainty information. However, in practice, optimizing this complex distribution for deep architectures leads to convergence issues (as mentioned before) such that the more simple distribution proposed by us leads to a more reliable approximation.

\subsection{Methodology}\label{theorie}

\begin{figure}[h!]
\centering
\includegraphics[width=0.45\textwidth]{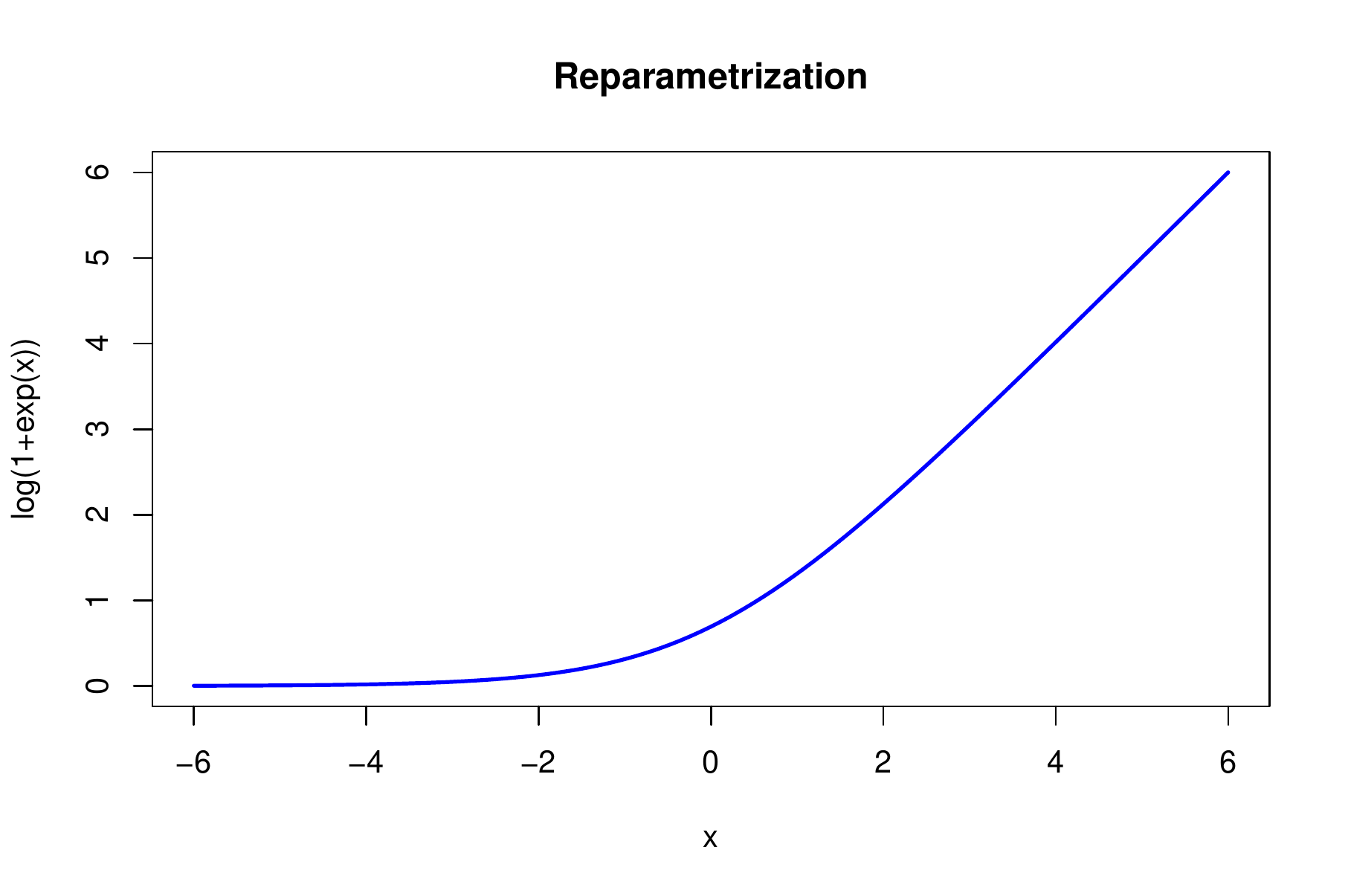}
\caption{Reparametrization of the uncertainty parameters.}
\label{reparam}
\end{figure}

Let $\underline{W}_i=(W_{i1},...,W_{iK_i})^T$ denote the random weights of the $i$-th network layer and further let \\$\underline{B}_i=(B_{i1},...,B_{ik_i})$ denote the corresponding random biases. In addition, let $\underline{\mathcal{E}}_i=(\mathcal{E}_{i1},...,\mathcal{E}_{iK_i})^T$ and $\underline{\mathcal{E}}_{bi}=(\mathcal{E}_{bi1},...,\mathcal{E}_{bik_i})^T$ be multivariate standard normal distributed. To set up the variational distribution, the random weights $\underline{W}_i$ and the random biases $\underline{B}_i$ are defined by
\begin{align}
\underline{W}_i &:=\underline{m}_i\odot(\underline{1}_{K_i} +\tau_i\underline{\mathcal{E}}_i) \label{eq:randW}\\
\tau_i &:=\operatorname{ln}(1+\operatorname{exp}(\delta_i)) \label{eq:tau_i}\\
\underline{B}_i &:=\underline{m}_{bi}\odot(\underline{1}_{k_i}+\tau_{bi}\underline{\mathcal{E}}_{bi}) \label{eq:randB}\\
\tau_{bi} &:=\operatorname{ln}(1+\operatorname{exp}(\delta_{bi})) \label{eq:tau_bi}
\end{align}
where $\underline{m}_i\in\mathbb{R}^{K_i}$, $\underline{m}_{bi}\in\mathbb{R}^{k_i}$, $\delta_i\in\mathbb{R}$ and $\delta_{bi}\in\mathbb{R}$ are variational parameters and $\odot$ denotes the Hadamard product, i.e. element-wise multiplication. This implies that $\underline{W}_i$ and $\underline{B}_i$ are multivariate normal distributed according to
\begin{align}
\underline{W}_i &\sim N\left(\underline{m}_i,\tau_i^2 \operatorname{diag}(\underline{m}_i)^2\right) \label{eq:weightGauss}\\
\underline{B}_i &\sim N(\underline{m}_{bi}, \tau_{bi}^2\operatorname{diag}(\underline{m}_{bi})^2). \label{eq:weightBias}
\end{align}
The reason why weights and biases are not directly defined by the Gaussians given in Eqs. \ref{eq:weightGauss} and \ref{eq:weightBias}, can be found in the network optimization. During optimization in each iteration, a sample is drawn from the random network parameters in order to adjust the variational parameters by gradient descent. A direct sampling from Eqs. \ref{eq:weightGauss} and \ref{eq:weightBias} would mask the variational parameters and therefore exclude them from optimization. The indirect sampling according to Eqs. \ref{eq:randW} and \ref{eq:randB} ensures that the variational parameters can be updated. In addition, $\tau_i$ and $\tau_{bi}$ are reparametrized according to Eqs. \ref{eq:tau_i} and \ref{eq:tau_bi} since they regulate the variance of the random network parameters and therefore should not take values less than zero. As one can see in \figurename \ref{reparam} the reparameterization  used guarantees that $\tau_i$ and $\tau_{bi}$ stay positive during optimization.
Finally we define the overall variational distribution by
$$q_{\underline{\phi}}(\underline{w})=\prod\limits_{i=1}^{d}q_{\underline{\phi}_i}(\underline{w}_i)q_{\underline{\phi}_{bi}}(\underline{b}_i)$$
$$\text{with}~~~~\underline{\phi}_i:=\{\underline{m}_i,\delta_i\},~
\underline{\phi}_{bi}:=\{\underline{m}_{bi},\delta_{bi}\}$$
by assuming that $\underline{W}_1,...,\underline{W}_d$ and $\underline{B}_1,...,\underline{B}_d$ are stochastically independent and with $q_{\underline{\phi}_i}(\underline{w}_i)$ and $q_{\underline{\phi}_{bi}}(\underline{b}_i)$ denoting densities of normal distributions according to Eqs. \ref{eq:weightGauss} and \ref{eq:weightBias}. The depth of the network is denoted by $d$. Thus, parameter uncertainty is expressed layer-wise and relative to the parameter expectation values. It is assumed that the uncertainty in network parameters depends on the layers they belong to and not on single parameters or neurons. In analogy to the variational density, the normal prior $p(\underline{w})$  is defined as
$$p(\underline{w}):=\prod\limits_{i=1}^{d}p(\underline{w}_i)p(\underline{b}_i)$$
$$\text{with}~~~\underline{W}_i\sim N(\underline{\mu}_i,\zeta_i^2 I_{K_i\times K_i})~~~$$ \\ $$\text{and}~~~\underline{b}_i\sim N(\underline{\mu}_{bi}, \zeta_{bi}^2I_{k_i\times k_i}).$$
Therefore, the a priori uncertainty is again expressed layer-wise. The fact that both the variational distribution and the prior factorize simplifies the calculation of the Kullback Leibler divergence between those two. It is given by the sum of the layer-wise divergences:
\begin{small}
\begin{align*}
&D_{KL}(q_{\underline{\phi}}(\underline{w})||p(\underline{w}))=\mathbb{E}_{q_{\underline{\phi}}(\underline{w})}\left(\operatorname{ln}\frac{\prod\limits_{i=1}^{d}q_{\underline{\phi}_{i}}(\underline{w}_i)q_{\underline{\phi}_{bi}}(\underline{b}_i)}{\prod\limits_{i=1}^{d}p(\underline{w}_i)p(\underline{b}_i)}\right)\\
&=\mathbb{E}_{q_{\underline{\phi}}(\underline{w})}\left\{\sum\limits_{i=1}^d\left[\operatorname{ln}(q_{\underline{\phi}_{i}}(\underline{w}_i))-\operatorname{ln}(p(\underline{w}_i))\right]\right.\\
&\left.+\sum\limits_{i=1}^d\left[\operatorname{ln}(q_{\underline{\phi}_{bi}}(\underline{b}_i))-\operatorname{ln}(p(\underline{b}_i))\right]\right\}\\
&=\sum\limits_{i=1}^d\left[\mathbb{E}_{q_{\underline{\phi}_i}(\underline{w}_i)}\operatorname{ln}\frac{q_{\underline{\phi}_i}(\underline{w}_i)}{p(\underline{w}_i)}\right]+\sum\limits_{i=1}^d\left[\mathbb{E}_{q_{\underline{\phi}_{bi}}(\underline{b}_i)}\operatorname{ln}\frac{q_{\underline{\phi}_{bi}}(\underline{b}_i)}{p(\underline{b}_i)}\right]\\
&=\sum\limits_{i=1}^d\left[D_{KL}(q_{\underline{\phi}_i}(\underline{w}_i)||p(\underline{w}_i))\right]+\sum\limits_{i=1}^d\left[D_{KL}(q_{\underline{\phi}_{bi}}(\underline{b}_i)||p(\underline{b}_i))\right]
\end{align*}
\end{small}
Given that the KL-divergence between two $p$-dimensional Gaussians $h(\underline{x})=N(\underline{x};\underline{\mu}_h,\Sigma_h)$ and $g(\underline{x})=N(\underline{x};\underline{\mu}_g,\Sigma_g)$ is given by \citep{kldivergence}
\begin{align*}
D_{KL}(h||g)=\frac{1}{2}&\left[\operatorname{ln}\frac{|\Sigma_g|}{|\Sigma_h|}+\operatorname{tr}(\Sigma_g^{-1}\Sigma_h)-p\right.\\
&\left.+(\underline{\mu}_h-\underline{\mu}_g)^T\Sigma_g^{-1}(\underline{\mu}_h-\underline{\mu}_g)\right]
\end{align*}
it is easy to calculate $D_{KL}(q_{\underline{\phi}}(\underline{w})||p(\underline{w}))$. Up to an additive constant, which plays no role in the optimization, it is given by:
\begin{small}
\begin{align*}
&D_{KL}(q_{\underline{\phi}}(\underline{w})||p(\underline{w}))\propto\\
&\frac{1}{2}\sum\limits_{i=1}^{d}\left[-\operatorname{ln}(\tau_i^{2K_i}|\operatorname{diag}(\underline{m}_i)^2|)+\frac{\tau_i^2}{\zeta_i^2}||\underline{m}_{i}||_2^2+\frac{1}{\zeta_i^2}||\underline{m}_i-\underline{\mu}_i||_2^2\right.\\
&\left.-\operatorname{ln}(\tau_{bi}^{2k_i}|\operatorname{diag}(\underline{m}_{bi})^2|)+\frac{\tau_{bi}^2}{\zeta_{bi}^2}||\underline{m}_{bi}||_2^2+\frac{1}{\zeta_{bi}^2}||\underline{m}_{bi}-\underline{\mu}_{bi}||_2^2\right]
\end{align*}
\end{small}

In order to train a network $\underline{f}$ in a Bayesian way according to our approach, i.e. with weights sampled from $q_{\underline{\phi}}(\underline{w})$ and $D_{KL}(q_{\underline{\phi}}(\underline{w})||p(\underline{w}))$ as regularization, the partial derivatives with respect to $m_{ij},m_{bij},\delta_{i}$ and $\delta_{bi}$ are needed. The derivatives of the neural net can easily be expressed in terms of the classical derivatives in non-Bayesian deep learning. According to the chain rule the derivatives with respect to $m_{ij}$ and $m_{bij}$ are given by
\begin{align}
\frac{\partial \underline{f}}{\partial m_{ij}}&=\frac{\partial \underline{f}}{\partial w_{ij}}\cdot\frac{\partial w_{ij}}{\partial m_{ij}}=\frac{\partial \underline{f}}{\partial w_{ij}}\cdot(1+\tau_{i}\varepsilon_{ij}) \label{eq:grad_weight}\\
\frac{\partial \underline{f}}{\partial m_{bij}}&=\frac{\partial \underline{f}}{\partial b_{ij}}\cdot\frac{\partial b_{ij}}{\partial m_{bij}}=\frac{\partial \underline{f}}{\partial b_{ij}}\cdot(1+\tau_{bi}\varepsilon_{bij}) \label{eq:grad_bias}
\end{align}
where $\frac{\partial \underline{f}}{\partial w_{ij}}$ and $\frac{\partial \underline{f}}{\partial b_{ij}}$ are calculated as in the non-Bayesain case. Note that
\begin{align*}
w_{ij}&=m_{ij}(1+\tau_{i}\varepsilon_{ij})\\
b_{ij}&=m_{bij}(1+\tau_{bi}\varepsilon_{bij})
\end{align*}
are denoting samples from $W_{ij}$ and $B_{ij}$, respectively.
Analogously, the derivatives with respect to $\delta_{i}$ and $\delta_{bi}$ are given by:
\begin{align}
\frac{\partial \underline{f}}{\partial \delta_{i}}&=\sum\limits_j\frac{\partial \underline{f}}{\partial w_{ij}}\cdot m_{ij}\varepsilon_{ij}\frac{e^{\delta_i}}{1+e^{\delta_i}} \label{eq:grad_deltai}\\
\frac{\partial \underline{f}}{\partial \delta_{bi}}&=\sum\limits_j\frac{\partial \underline{f}}{\partial b_{ij}}\cdot m_{bij}\varepsilon_{bij}\frac{e^{\delta_{bi}}}{1+e^{\delta_{bi}}} \label{eq:grad_deltabi}
\end{align}
Further, one can easily verify that the partial derivatives of the KL-divergence $D_{KL}$ are given by:
\begin{align}
\frac{\partial}{\partial m_{ij}}D_{KL}&=-\frac{1}{m_{ij}}+\frac{\tau_i^2}{\zeta_i^2}m_{ij}+\frac{1}{\zeta_i^2}(m_{ij}-\mu_{ij}) \label{eq:DKL_weights}\\
\frac{\partial}{\partial \delta_{i}}D_{KL}&=\frac{\operatorname{exp}(\delta_i)}{1+\operatorname{exp}(\delta_i)}\left[\frac{\tau_i}{\zeta_i^2}||\underline{m}_i||_2^2-\frac{K_i}{\tau_i}\right]\\
\frac{\partial}{\partial m_{bij}}D_{KL}&=-\frac{1}{m_{bij}}+\frac{\tau_{bi}^2}{\zeta_{bi}^2}m_{bij}+\frac{1}{\zeta_{bi}^2}(m_{bij}-\mu_{bij}) \label{eq:DKL_bias}\\
\frac{\partial}{\partial \delta_{bi}}D_{KL}&=\frac{\operatorname{exp}(\delta_{bi})}{1+\operatorname{exp}(\delta_{bi})}\left[\frac{\tau_{bi}}{\zeta_{bi}^2}||\underline{m}_{bi}||_2^2-\frac{k_i}{\tau_{bi}}\right] \label{eq:DKL_deltabi}
\end{align}

\subsection{Implementation}

We implemented the approach illustrated above by modifying the popular open-source Caffe framework \citep{jia2014caffe, caffe-github}. For the convolutional layer and the inner product layer, the layer parameter "blobs" were extended to include the additional variance terms for the weights and biases, $\delta_i$ and $\delta_{bi}$ respectively, as well as the current realizations of $\underline{\mathcal{E}}_i$ and $\underline{\mathcal{E}}_{bi}$. The weights and biases of the classical imlpementation are here interpreted as the variational parameters $\underline{m}_i$ and $\underline{m}_{bi}$. In addition, for each layer, static arrays to hold the prior expectation values for the weights and biases $\mu_{ij}$ and $\mu_{bij}$ and the corresponding a priori variance terms $\zeta_i$ and $\zeta_{bi}$ were introduced. The variances and expectation values of the prior distributions and the starting values for the variances and expectation values of the variational distributions $\delta_i$ and $\delta_{bi}$ can be set for each layer in the network definiton prototext file. During each forward pass, one sample is drawn from the $\underline{\mathcal{E}}_i$ and the $\underline{\mathcal{E}}_{bi}$. From this, the random weights and biases used in the forward pass, $\underline{W}_i$ and $\underline{B}_{i}$, are calculated according to Eqs. \ref{eq:randW}-\ref{eq:tau_bi} using the current variational parameters and the current realizations of $\underline{\mathcal{E}}_i$ and $\underline{\mathcal{E}}_{bi}$. During the backward pass, the gradients of the weights and biases are adapted according to Eqs. \ref{eq:grad_weight} and \ref{eq:grad_bias}. In addition, the gradients for the new variance parameters are calculated according to Eqs. \ref{eq:grad_deltai} and \ref{eq:grad_deltabi}. For each gradient, the additional term due to the Kullback-Leibler divergence is added following Eqs. \ref{eq:DKL_weights}-\ref{eq:DKL_deltabi}. These gradients are then used to calculate the updated variational parameters and thus the updated weights and biases to be used for the next forward pass. The additional term to the loss function due to the Kullback-Leibler divergence was not computed as it is not needed for the optimization. 

\section{Experimental Results} \label{experi}
%short description MNIST
%detailed description LeNet

In this section, the proposed approach is tested. Basis of all experiments is the benchmark dataset MNIST \citep{mnist} together with the architecture LeNet \citep{lenet}. The MNIST dataset consists of $70,000$ images of handwritten digits, from which $60,000$ build up the training dataset and the remaining $10,000$ build up the testing data. The specific version of LeNet used is the same described in \cite{Gal:conv}. Therefore the first convolutional layer generates $20$ feature maps, while the second one extracts $50$ features. Both layers use $(5\times 5)$ kernels. Max-pooling with kernel size $(2\times 2)$ and stride $2$ is applied after both convolutional layers. The first fully connected layer consists of $500$ neurons, the second one covers only $10$ since there are $10$ different digits. Moreover, the first fully connected layer uses the rectified linear unit as activation and the other ones the identity function.

\subsection{Frequentist Models}\label{frequ_train}
%without dropout
%with dropout
%with dropout and exchnaged training and teting data

%to compare results with bayesain analogues
%one description of optimizer
%one example of learning curve and so on
%one table with all accuracies

In order to get an idea how well our Bayesian approach performs it should be compared to the classical, i.e. the frequentist, approach. Therefore, LeNet is trained three times in the classical way. First, without dropout, then with dropout \textit{(dropping rate $0.5$)} applied after the first inner product layer, and finally, with dropout applied as before and exchanged training and testing datasets. Exchanging training and testing data results in a significant reduction of the training data from $60,000$ to $10,000$ and should give an intuition how well bayesian models work for limited training data.

\begin{figure}[h!]
\centering
\includegraphics[width=0.45\textwidth]{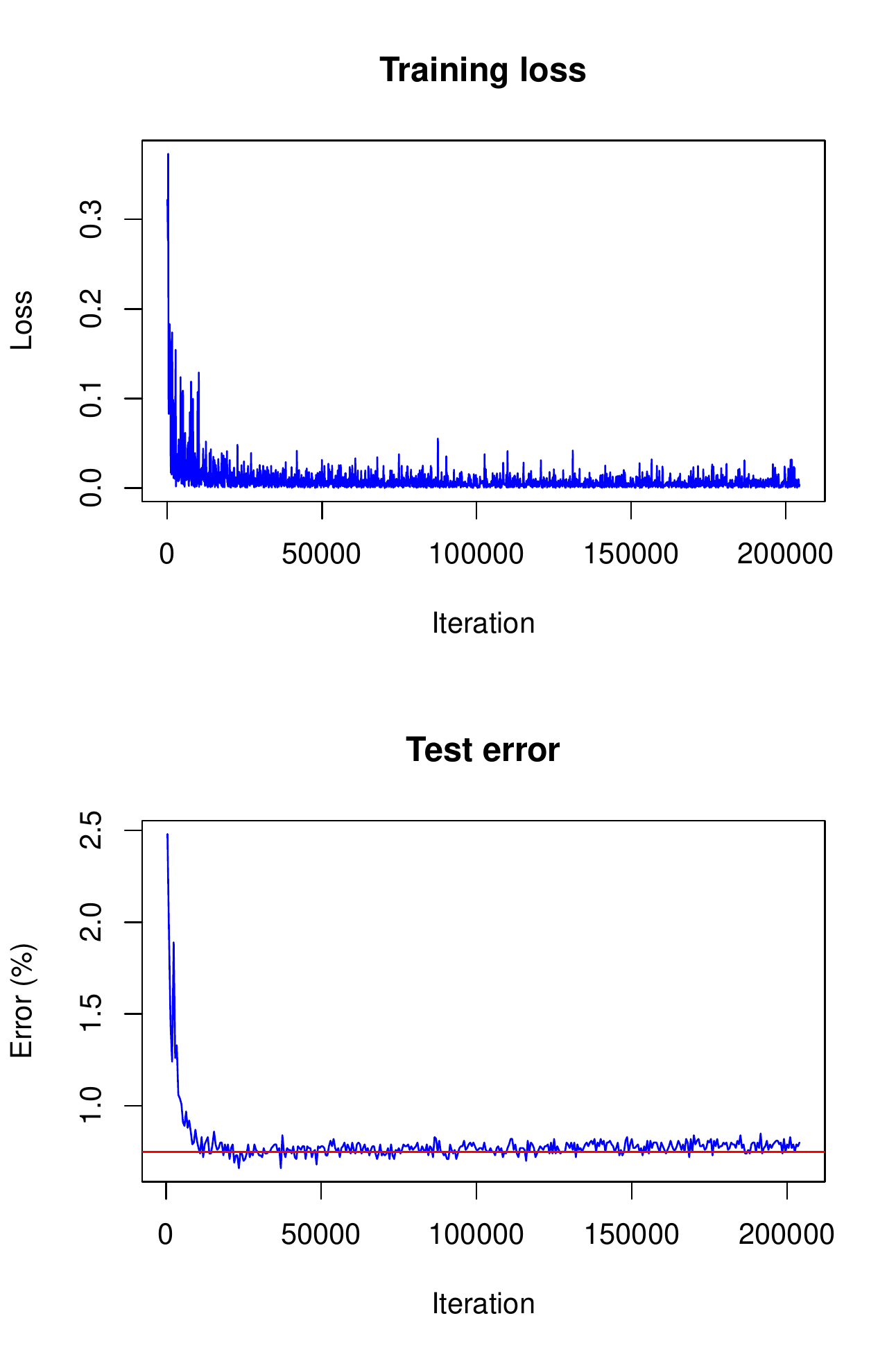}
\caption{Training visualization of frequentist LeNet with dropout. The horizontal line marks the achieved accuracy.}
\label{frequentistic_accuracies}
\end{figure}

All three models are optimized the same way. To prevent overfitting, the Euclidean norm of the network weights is penalized with a factor of $0.0005$. As usual in deep learning, the optimization procedure applied is mini-batch gradient descent. A batch size of $64$ is chosen. The learning rate used in the $i$-th iteration is given by $0.01*(1+0.0001*i)^{-0.75}$. Momentum is used and set to $0.9$. The accuracies achieved are given in Table \ref{frequentist_accuracies} and are expressed by the corresponding test error.
\begin{table}[h!]
\renewcommand{\arraystretch}{1.3}
\caption{Accuracies frequentist LeNet}
\label{frequentist_accuracies}
\centering
\begin{tabular}{|l|l|}
\hline
model &  test error \\
\hline
without dropout	   & 0.9\%\\
with dropout       & 0.75\%\\
with dropout and exchanged data      & 1.94\%\\
\hline
\end{tabular}
\end{table}
The training converged quite similar in all three cases. A visualization of the training loss and test error for the second model, i.e. the model trained with dropout, is shown in \figurename \ref{frequentistic_accuracies}. This figure will serve for comparison of the Bayesian and the frequentist training process.

\subsection{Bayesian Models}\label{bayes_models}

%without dropout
%with dropout
%with dropout and exhanged training and testing data

%one description of prior + posterior initialization

%one example of learning curve and so on
%heavy fluctuations -> maybe averaging more samples before updating
%one table with all uncertaintys
%one table with all accuracies

To test and verify our Bayesian approach, LeNet is trained three times with the MNIST dataset. In analogy to the frequentisitc training (see section Section \ref{frequ_train}), LeNet is trained first without dropout, then with dropout, and finally, with dropout and exchanged training and testing data. In contrast to Gal and Ghahramani, we interpret dropout training as simultaneous training of multiple Bayesian models and assume that combining multiple models will result in a better accuracy than using just one model. Thus, during testing, the weight scaling inference rule, which states that each neuron should be used but multiplied with the dropping ratio, is not applied. Rather in the testing phase, neurons are randomly dropped in order to sample from the set of simultaneously trained Bayesian models and combine their predictions to one overall prediction.

In contrast to the non-Bayesian case, a penalization of the Euclidean norm does not take place since in the Bayesian case deviations from the a priori distribution are penalized. As there is not really a priori information available, the prior is used to express the wish that values should not diverge. Thus, the a priori expectation value is specified as zero for all network parameters, and further, the a priori standard deviation is chosen to be $5$ for all weights and $10$ for all biases. The variance for the biases is chosen to be larger since biases act on linear combinations of neuron outputs with network weights as coefficients and therefore may take on larger values.
\begin{figure}[h!]
\centering
\includegraphics[width=0.45\textwidth]{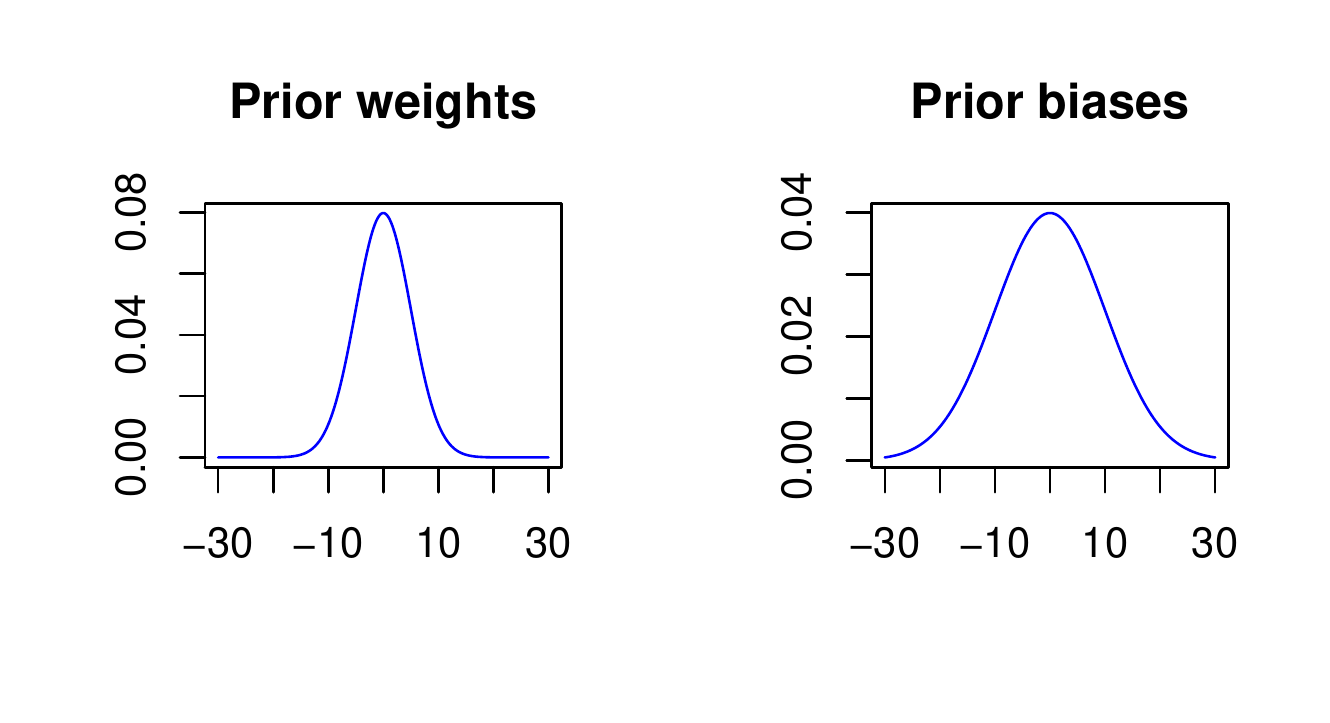}
\caption{Visualization of the a priori distribution. Bias terms may take on larger values since they act on sums.}
\label{Prior}
\end{figure}
It should be mentioned that the penalization strength of the KL-divergence between the variational distribution and the a priori distribution is chosen smaller than recommended in the theoretical considerations in Section \ref{theorie} because of convergence problems. Empirically, we found that we have to scale the penalization strength down by a factor of $100$ to ensure convergence. While somewhat puzzling, this does not matter since there is not really a priori information available and the network parameters took small values in all experiments even with the reduced penalization. It should also be mentioned, that our implementation easily lends itself to 'Bayesian transfer learning' in analogy to classical transfer learning \citep{bengioJMLR2012}, where the results of a previous training run with a large dataset are optimized for a more targeted application by finetuning the network with a smaller but specific training dataset. In the Bayesian case, the information about the posterior distribution of the network parameters in the pre-trained network will then be used to specify a prior for the finetuning step. This is subject to future work. 

\begin{figure}[h!]
\centering
\includegraphics[width=0.45\textwidth]{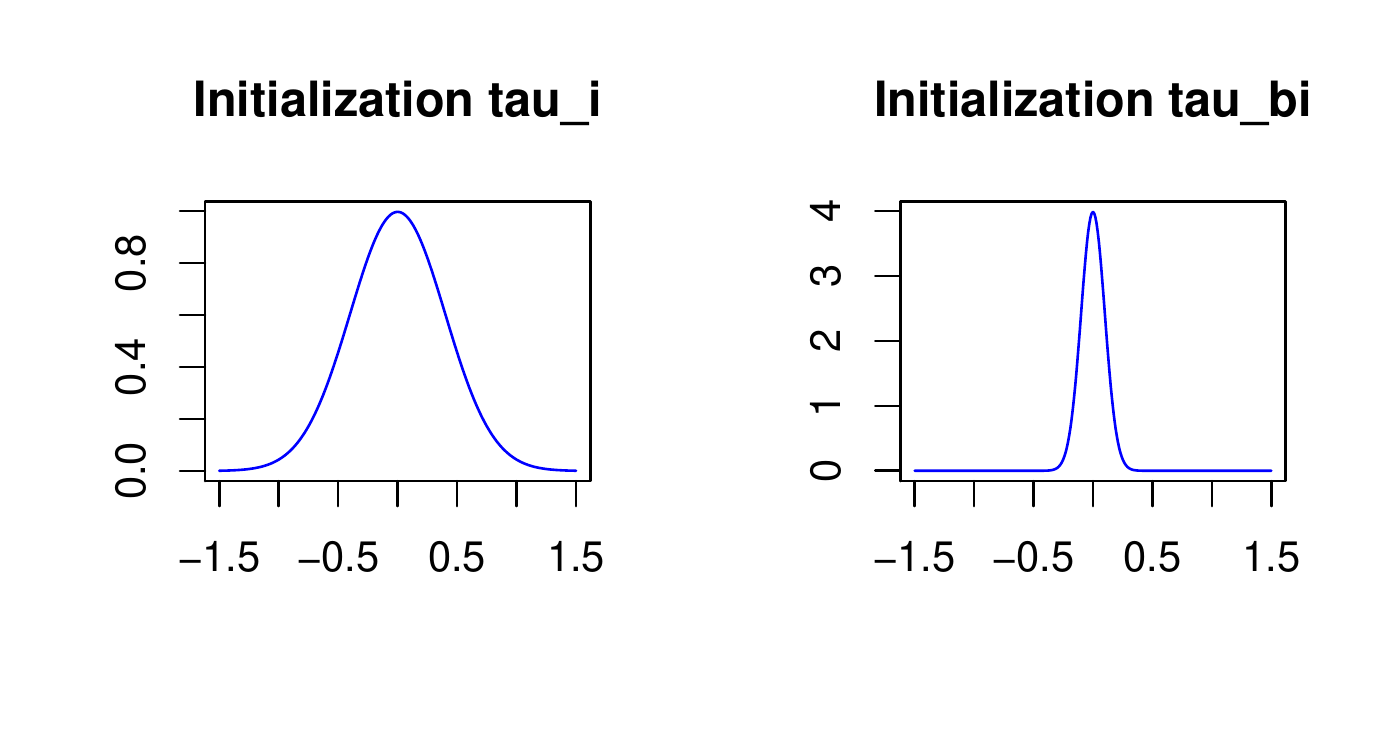}
\caption{Visualization of the initialization of $\tau_i$ and $\tau_{bi}$. Network weights can differ at most by the size of their expectation value from their expectation value.}
\label{initialization}
\end{figure}

In order for the Bayesian networks to converge, the parameters $\tau_{i}$ and $\tau_{bi}$ \textit{(see Section \ref{theorie})} which specify the a posteriori uncertainty have to be initialized carefully. Therefore, $\tau_{i}$ is initialized as $0.4$ and $\tau_{bi}$ as $0.1$ in all network layers except for the first fully connected one which is treated separately. In this initialization, neural net weights can differ at most by the size of their expectation value from their expectation value, see \figurename \ref{initialization}.
This is a reasonable way to start since stronger deviations from the expectation values would mean that weights are even unsure about their algebraic sign, which might lead to convergence issues, if assumed for a majority of the network layers. In addition, assuming biases to vary less is not unusual since there are relatively few of them and they have a strong influence on the model since they act on sums. The reason why the first fully connected layer is treated differently is that it covers much more parameters than the other layers. Indeed it includes $400,000$ weights, while all the other layers together only contain $30,500$ weights. Due to the large number of parameters in the first fully connected layer, we assume that the model will be more uncertain in the network parameters of this layer. So $\tau_{i}$ is initialized with $1$ and $\tau_{bi}$ with $0.2$.

Finally, all Bayesian models are optimized with the same mini-batch optimization procedure as their frequentist analogues. For computing the model accuracies, each test example is propagated $100$ times through the network using Caffe's bindings to Python. The test errors achieved and the absolute and relative decreases in the error with respect to the non-Bayesian models are given in Table \ref{bayes_accuracies}.
\begin{table}[h!]
\renewcommand{\arraystretch}{1.3}
\caption{Accuracies Bayesian LeNet}
\label{bayes_accuracies}
\centering
\begin{tabular}{|l|l|l|l|}
\hline
model &  test error  & abs. decr. & rel. decr.\\
\hline
without dropout	   				 & 0.85\% 	& 0.05	&5.6\%\\
with dropout      				 & 0.71\%	& 0.04	&5.3\%\\
dropout \& exch. data  			 & 1.64\% 	& 0.3	&15.5\%\\
\hline
\end{tabular}
\end{table}
One can see that the Bayesian models always perform better than their frequentist analogues. For the first two models the accuracy is only slightly better, while the third model shows a significant improvement, especially if one considers the relative decrease of the test error. It is not surprising that the increase in accuracy is only small for the first two models since all models considered converge very well and do not suffer from overfitting because there is plenty of training data available. The third model which is trained using only $10,000$ images shows signs of overfitting in the non-Bayesian case. The Bayesian network however is more robust towards overfitting and thus performs significantly better. This illustrates the advantage of Bayesian deep learning in the presence of only a limited number of training images. 

\begin{figure}[h!]
\centering
\includegraphics[width=0.4\textwidth]{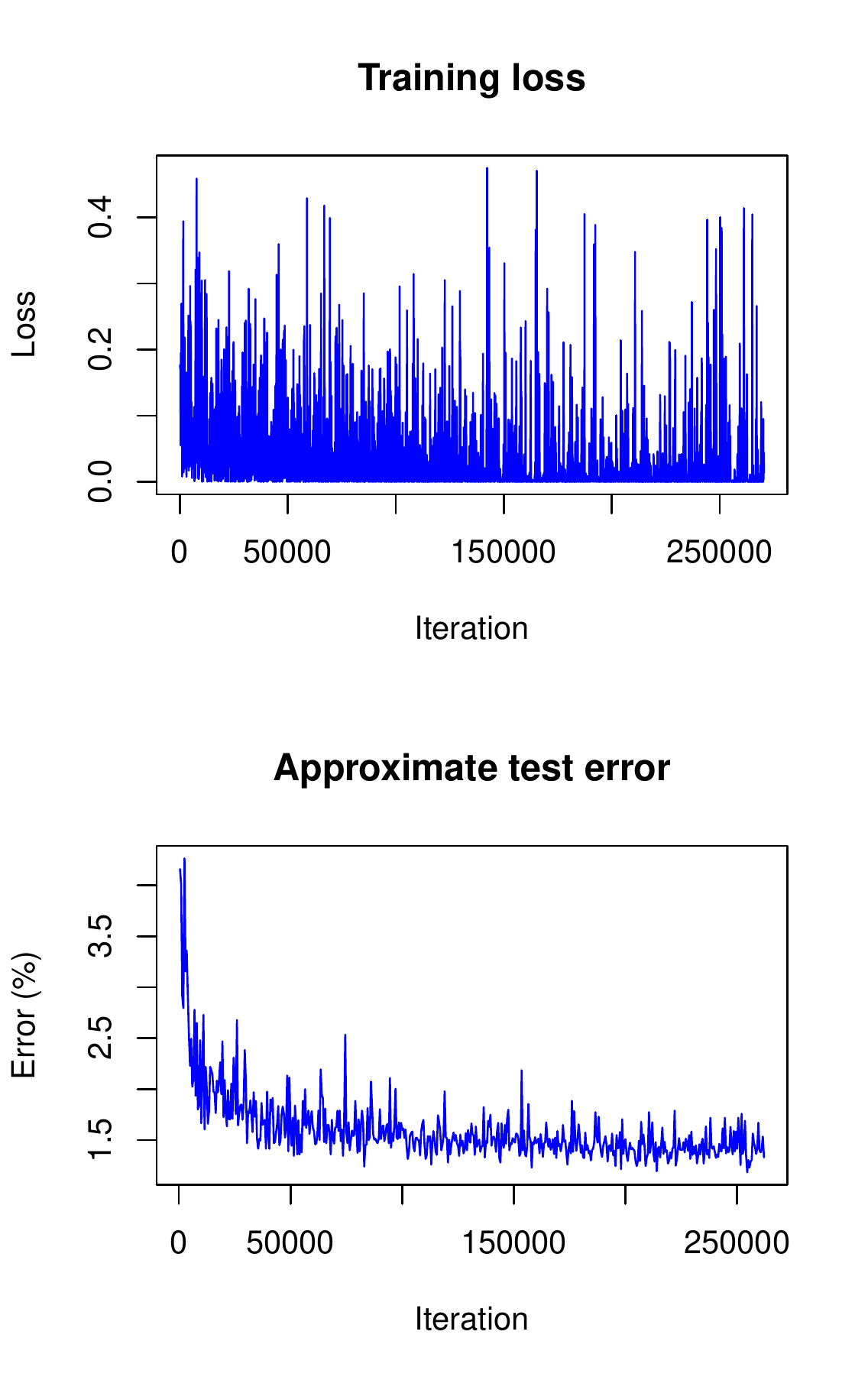}
\caption{Visualization of the network training without dropout.}
\label{training_bayes}
\end{figure}

\begin{figure}[h!]
\centering
\includegraphics[width=0.4\textwidth]{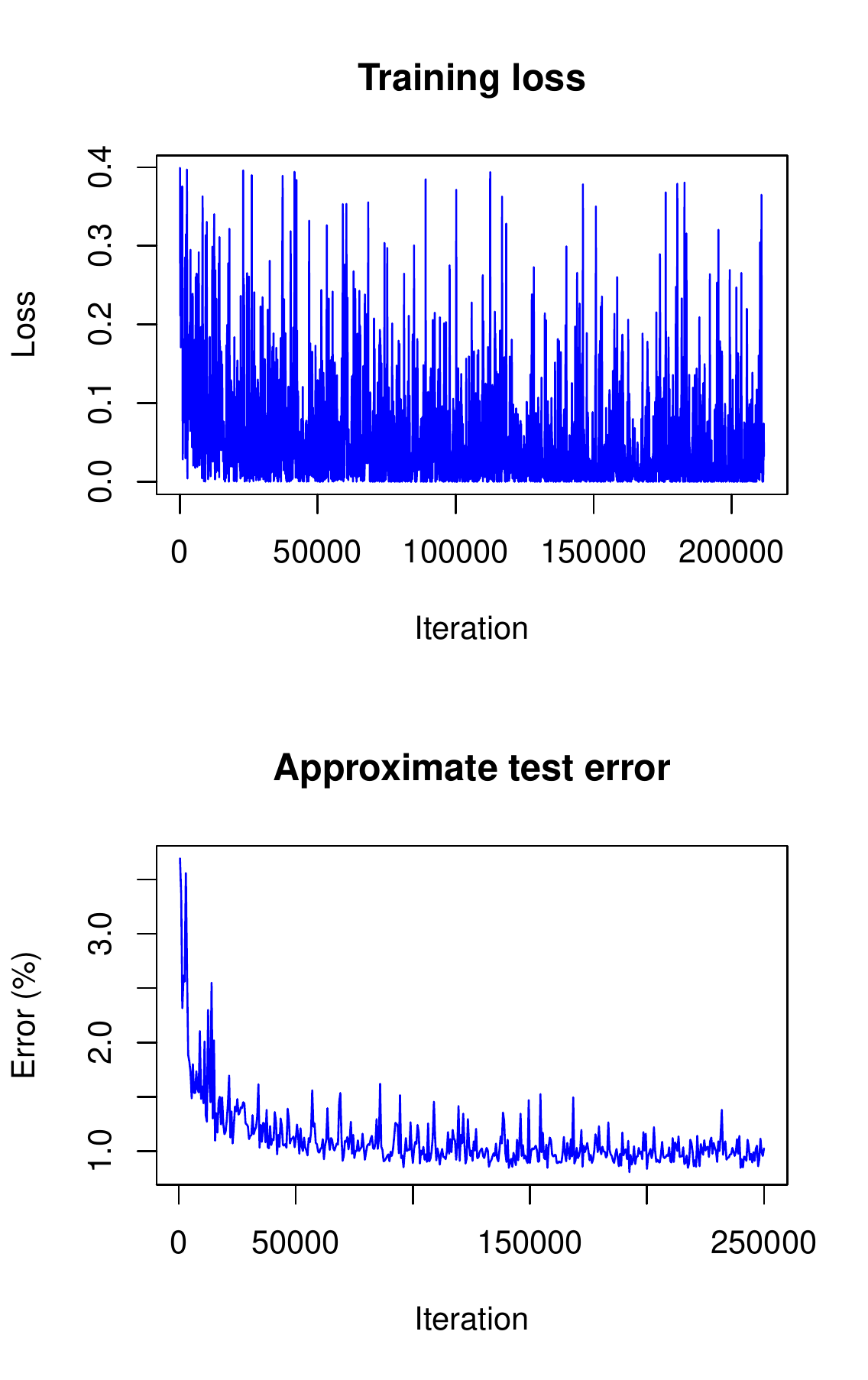}
\caption{Visualization of the network training with dropout.}
\label{training_drop_bayes}
\end{figure}

It is interesting to see how LeNet converges following our Bayesian approach. In \figurename \ref{training_bayes} and \figurename \ref{training_drop_bayes} the training is visualized for the first and the second model, i.e. the model trained without dropout and the model trained with dropout.
In contrast to the frequentist case, only the approximate test error is plotted. This means that only one sample of each testing image is used for predictions and that the weight scaling inference rule is applied. Currently, the Caffe framework does not provide other options for the testing phase during optimization. Nonetheless, the imprecise approximation of the test error gives a rough estimate of the real test error and therefore helps to understand what happens with the model accuracy during training. One can see that the loss \textit{(plotted without the term due to the KL-divergence)} fluctuates heavily during training due to the random samples drawn from the variational distribution. However, the test error decreases quickly as in the non-Bayesian case and seems to keep decreasing as training goes on. This is not the case for the frequentist model (see \figurename \ref{frequentistic_accuracies}) for which the test error seems to increase slowly. This, again, indicates the strength of our approach against overfitting.
\begin{figure*}[h!]
\centering
\includegraphics[width=0.95\textwidth]{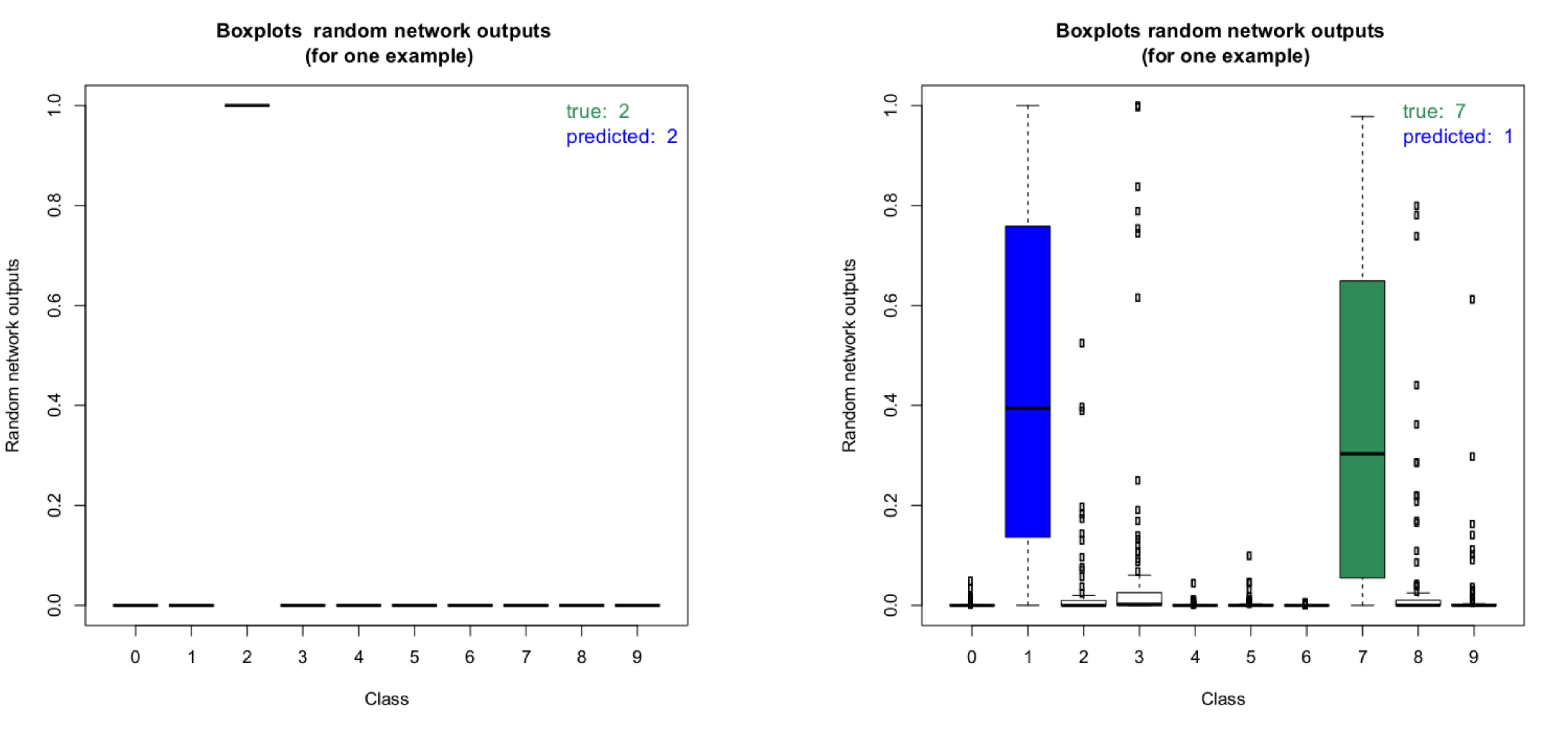}
\caption{Boxplots of the random network outputs of two representative images of the MNIST dataset. Left: Boxplot of a correct classification, right: boxplot of an incorrect classification result.}
\label{fig:boxplots}
\end{figure*}

The a posteriori uncertainties are quite the same for all three Bayesian models. In Table \ref{posterior_lenet_drop}, the uncertainties for the second model, i.e. the model trained with dropout, are given. One can see that the model uncertainty is small for all layers except for the first fully connected one.
\begin{table}[h!]
\renewcommand{\arraystretch}{1.3}
\caption{A posteriori uncertainty}
\label{posterior_lenet_drop}
\centering
\begin{tabular}{|l|l|l|}
\hline
layer & $\tau_i$ & $\tau_{bi}$\\
\hline
convolutional 1		& 0.003905073 & 0.01703058\\
convolutional 2	    & 0.000045391 & 0.1021243\\
fully connected 1   & 0.7580626   & 0.1471348\\
fully connected 2	& 0.02509901  & 0.00004402\\
\hline
\end{tabular}
\end{table}

\subsection{Reduced Model}
%baysian and non-Baysian
%accuracies
%uncertainty tables
%training visualization
In Section \ref{bayes_models} LeNet was trained in a Bayesian way \textit{(with dropout)}, resulting in a high model uncertainty of the first fully connected Layer. A value of $0.76$ for $\tau_3$ indicates that the network is not even sure about the algebraic sign of the weights in this layer. Therefore, we assume that the network architecture is not optimal and reduce the number of output neurons for the first fully connected layer from $500$ to $250$. In the Bayesian case, this does not lead to a significant increase of the network accuracy but the network uncertainty for the first fully connected layer decreases significantly   as one can see in Table \ref{posterior_lenet_drop_red}.
\begin{table}[h!]
\renewcommand{\arraystretch}{1.3}
\caption{A posteriori uncertainty reduced model}
\label{posterior_lenet_drop_red}
\centering
\begin{tabular}{|l|l|l|}
\hline
layer & $\tau_i$ & $\tau_{bi}$\\
\hline
convolutional 1		& 0.003570068 & 0.01361556\\
convolutional 2	    & 0.000045395 & 0.1025237\\
fully connected 1   & 0.5672782   & 0.1359651\\
fully connected 2	& 0.01789308  & 0.00004324\\
\hline
\end{tabular}
\end{table}
This result indicates that the Bayesian approach can be used to optimize the model architecture both in terms of accuracy and model size for a given training and testing dataset. In this particular case, we were able to reduce the number of parameters by almost a factor of $2$ while achieving the same accuracy. Even more interesing, when the reduced model is trained the classical way, the achieved accuracies become as good as for the Bayesian model, indicating again that the initial model was suffering from overfitting.

\subsection{Prediction uncertainty}
\begin{figure*}[h!]
\centering
\includegraphics[width=0.95\textwidth]{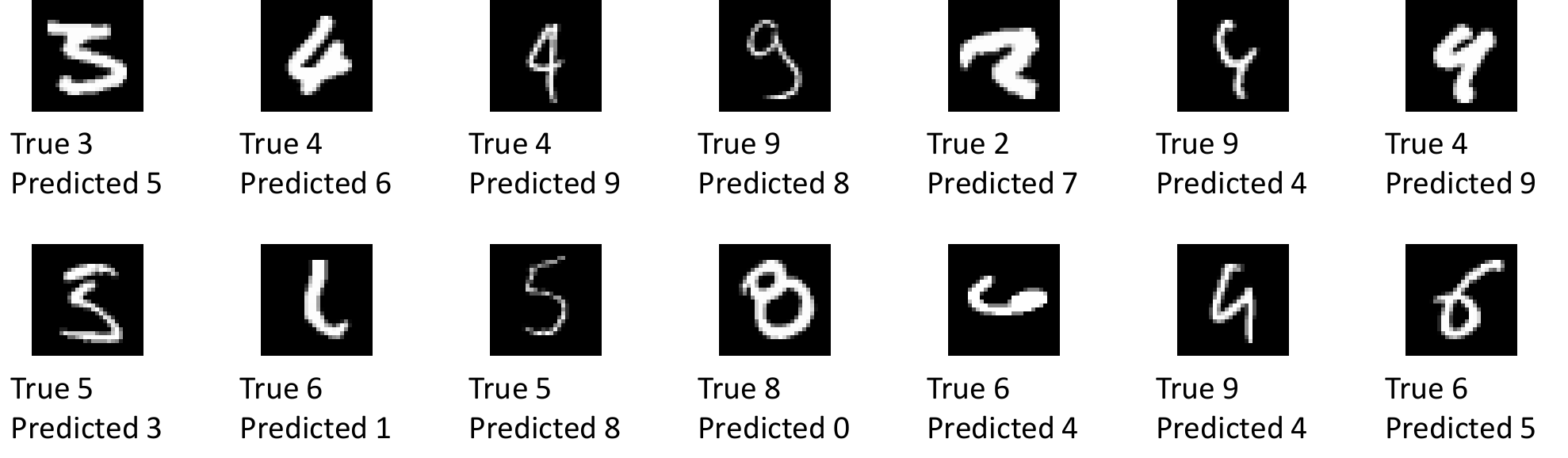}
\caption{All 14 images where the network was sure within 95\% credible intervals about its wrong classification result.}
\label{fig:wrong-sure}
\end{figure*}
In addition to providing information about the model uncertainty, our approach can also be used to determine the uncertainties of the predictions. Due to the random sampling of the weights and biases during each forward pass, accurate credible intervals can be estimated by performing multiple forward passes  per image. 
This information can be used in applications using our algorithm for classification. For example, a check for statistical significance for the classification result can be performed and the result can be used to decide about the next steps in the application (e.g. proceed autonomously, repeat classification, escalate to user, etc.).  Figure~\ref{fig:boxplots} shows two boxplots of the random network outputs (model without dropout) for two representative images from the MNIST test data set. On the left, the boxplot for an image with correct classification result is shown. Clearly, the network is very certain about this classification result. On the right, the boxplot for an image with wrong classification result is shown. As can be seen, the result for the wrongly predicted label is not statistically significant as there is a clear overlap between the boxes of the true label and the predicted one. These boxplots were computed by performing the inference 100 times for each image. It is interesting to note that in the case of the wrongly predicted image on the right, the network produces very high outlier probabilities for other classes besides the true and the predicted label. This illustrates the potential for deterministic networks to produce wrong classifications with very high class probabilities. Checking for all images if the estimated 95\% credible intervals of the predicted classes overlap with the 95\% intervals of the other classes gives further insight into the prediction capabilities of the network. Table~\ref{tab:results_all} summarizes the results for the model without dropout, see section~\ref{bayes_models}.
\begin{table}[h!]
\renewcommand{\arraystretch}{1.3}
\caption{Summary of  network performance (network without dropout)}
\label{tab:results_all}
\centering
\begin{tabular}{|l|l|l|}
\hline
 & quite certain & uncertain \\
\hline
correct    & 9609  & 297 \\
wrong	    & 14 & 80 \\
\hline
\end{tabular}
\end{table}
As can be seen, the overwhelming majority of classification results is correct and the network is also confident about these predictions. About 300 images are classified correctly, but the network is not sure within 95\% credible intervals. A total of 94 images are classified incorrectly. In the vast majority of these cases, the network is unsure about the classification result. In only 14 cases, the network is quite sure about its wrong classification. Please note, that due to the random sampling of network parameters, the results are slightly different each time they are computed unless a very large number of forward passes is performed for each image. This is also the reason why the number of missclassified images in this section differs from the one obtained in section~\ref{bayes_models}. The uncertainty analysis presented here was performed separately. From an application point of view, the latter case is the most critical. Figure~\ref{fig:wrong-sure} shows all of the 14 images which have been classified wrongly with confidence by the network.   
More than half of these images visually resemble the predicted label at least as much as they resemble the true label. The remaining images are without a doubt wrongly classified. Some of these images can be excluded by raising the confidence level requirement. A detailed investigation into wrong yet confident classification results is left for further study. 
\section{Conclusion}
We present here for the first time a Bayesian approach to deep learning that allows for accurate calculation of the uncertainty of network predictions as well as the uncertainty of the model parameters while introducing only few additional parameters to be optimized. In particular, we introduce two variance terms per layer (one for the weight parameters, one for the biases) that are optimized during training along with the other network parameters. Compared to classical, frequentist models, our approach is more robust against overfitting. Especially for small training datasets, a significant improvement in accuracy is obtained with our approach. In addition, information about network uncertainty can be readily interpreted and used for improvements of network architecture. Finally, our approach provides accurate uncertainty information about the predictions of the network with potentially significant impact for real world applications.

%% For one-column wide figures use
%\begin{figure}
%% Use the relevant command to insert your figure file.
%% For example, with the graphicx package use
%  \includegraphics{example.eps}
%\end{figure}
%%
%% For two-column wide figures use
%\begin{figure*}
%% Use the relevant command to insert your figure file.
%% For example, with the graphicx package use
%  \includegraphics[width=0.75\textwidth]{example.eps}
%\end{figure*}
%\begin{table}
%\begin{tabular}{lll}
%\hline\noalign{\smallskip}
%\end{tabular}
%\end{table}

\section{Conflict of interest}
The authors declare that they have no conflict of interest.

\bibliographystyle{spbasic}      % basic style, author-year citations
\bibliography{Literatur}   % name your BibTeX data base

%\begin{thebibliography}{}
% etc
%\end{thebibliography}

\end{document}